\documentclass[conference]{IEEEtran}
\IEEEoverridecommandlockouts
\usepackage{booktabs}
\usepackage{multirow}
\usepackage{cite}
\usepackage{amsmath,amssymb,amsfonts}
\usepackage{algorithmic}
\usepackage{graphicx}
\usepackage{textcomp}
\usepackage{xcolor}
\usepackage{tabularx}
\usepackage{array}
\usepackage{siunitx}
\def\BibTeX{{\rm B\kern-.05em{\sc i\kern-.025em b}\kern-.08em
    T\kern-.1667em\lower.7ex\hbox{E}\kern-.125emX}}

\begin{document}

\title{EASTAR: Event-Anchored Semantic and Task-Adaptive Region Representation Learning}

\author{
\IEEEauthorblockN{Peng Liang$^{1\dagger}$, Genan Dai$^{2\dagger}$ and Bowen Zhang$^{2*}$}
\IEEEauthorblockA{\textsuperscript{1}College of Applied Science, Shenzhen University, Shenzhen, China}
\IEEEauthorblockA{\textsuperscript{2}School of Artificial Intelligence, Shenzhen Technology University, Shenzhen, China}
\IEEEauthorblockA{liangpeng\_lip@outlook.com, daigenan@sztu.edu.cn, zhang\_bo\_wen@foxmail.com}

\thanks{$^{\dagger}$~These authors contributed equally to this work.}
\thanks{$^{*}$~Corresponding author.}
}
\maketitle

\begin{abstract}
Urban region representation learning supports a wide range of urban analytics tasks by encoding regional functions and interactions, with human mobility serving as a central signal because it directly captures inter-region interactions. However, most mobility-based methods learn from aggregated origin-destination matrices, where trips driven by different activity purposes are mixed into purpose-agnostic interaction patterns. Although static urban attributes such as POIs and land use can enrich region-level semantics, they do not explicitly ground mobility interactions in activity-purpose signals. As a result, downstream adaptation is often performed over entangled embeddings, making it difficult to select task-relevant functional semantics.
To address these limitations, we propose EASTAR, an event-anchored and task-adaptive framework for urban region representation learning. EASTAR treats offline activity events as sparse semantic anchors and uses them to factorize aggregated mobility representations into multiple purpose-related semantic channels. It further introduces free channels to capture latent mobility patterns not covered by event data and a global channel to preserve overall spatial structure. To adapt the learned channel space to different downstream objectives, EASTAR designs a task-adaptive prompt router that initializes task-channel relevance from LLM-derived priors and refines the routing weights through downstream supervision. Experiments on two real-world urban datasets across four prediction tasks show that EASTAR consistently outperforms representative baselines, achieving up to 39.15\% reduction in MAE. Ablation and interpretability analyses further demonstrate the effectiveness of event anchoring, semantic channel construction, and task-adaptive routing.
\end{abstract}


\section{Introduction}

Urban region representation learning maps geospatial units into low-dimensional dense vectors that encode the functional attributes, spatial semantics, and interactions of urban regions~\cite{zheng2014urban}. These representations support a wide range of downstream urban analytics tasks, such as crime prediction~\cite{li2022spatial}, check-in forecasting~\cite{mvure}, and population estimation~\cite{urbanvlp}. As urban sensing data become increasingly rich, multi-view fusion has become the dominant paradigm~\cite{hafusion, urbanmmcl, mgfn}. It treats different urban data sources as complementary observational views of a region and fuses them into a unified urban representation.

Across these methods, human mobility consistently ranks among the strongest predictive signals in region representation learning~\cite{zemob, mobiclr, hgaurban, comprehensive}. For example, MGFN~\cite{mgfn}, relying on taxi flows alone, surpasses MVURE~\cite{mvure} by 10.0\% and 24.5\% in R² on check-in prediction in New York and Chicago, although MVURE also uses points of interest (POIs) and check-in data, and broader systematic comparisons report the same pattern~\cite{hafusion, comprehensive}. This is no coincidence: a region's functional role is not determined by its own attributes in isolation but emerges from its interaction patterns with other regions, consistent with spatial interaction theory~\cite{wilson1967statistical}, which holds that the functional organization of urban space is realized and sustained through inter-region flows. In contrast, POIs and remote sensing mainly characterize region-internal attributes and only indirectly reflect inter-region dependencies~\cite{hgi, tile2vec, URG}. Mobility therefore provides the most direct signal of inter-region interaction structure available in current multi-view systems~\cite{comprehensive}.

This very strength, however, exposes a critical limitation. Urban activity concerns not only where and when it happens, but also for what purpose~\cite{adamove}: trips for commuting, shopping, professional exchange, and social entertainment follow different spatial distributions and correspond to different urban functions~\cite{zhang2023regions,cgap}. An aggregated mobility matrix preserves the structural traces of these movements, such as origin-destination links and flow intensity, but largely obscures their underlying purposes~\cite{STSimSiam, UniMob}.

We therefore characterize inter-region functional interaction along two complementary dimensions. The structural dimension concerns which regions interact, how strongly, and how flows are distributed over space and time; the semantic dimension concerns why people interact and what activities drive the flow. Existing methods, whether they mine mobility structure in depth~\cite{mgfn}, disentangle spatio-temporal patterns~\cite{CrossSTAE}, or fuse multiple views~\cite{hafusion}, advance mainly along the structural dimension. Even when POIs are incorporated, they describe the facility composition of a region, such as how many restaurants or offices it contains~\cite{hgi, remvc}, rather than what people actually do there, so late fusion of POIs with mobility still fails to supply interaction-level semantics.

A real urban example illustrates this gap. In Manhattan, the Flatiron District concentrates technology companies, startup spaces, and professional community events, while the East Village is known for dining, bars, live music, and nightlife. In the evening, both can attract heavy cross-region inflows and show similar inflow intensity or temporal patterns in aggregated mobility. Yet the purposes behind these flows differ. The former is more likely tied to technical exchange, professional networking, and innovation, while the latter is more likely tied to dining, entertainment, and social leisure. Methods that rely only on aggregated mobility structure tend to treat such regions as similar, with limited ability to capture why these interactions occur~\cite{hdge,mobiclr}. This is a typical case of regions that are structurally similar but semantically different~\cite{SemiGTX}.

Supplying the semantic dimension requires signals that provide explicit, though sparse, evidence about activity purpose.
Event-based offline activity data has this property: it records not what facilities a region contains, but what activities are organized and attended there, each with a category, time, and location. We study this through event data from Meetup~\cite{meetup}, where every event carries an activity-category label, spatiotemporal information, and participant counts. Unlike approaches that attach activity  signals as an additional fusion view~\cite{mvure}, 
we treat it as a semantic anchor, a partial but explicit reference to activity purpose, and use it to guide the semantic factorization of aggregated mobility representations into purpose-related channels.
This distinguishes our approach from prior disentanglement, which separates signals across heterogeneous views rather than within the flow.

Grounding mobility representations with activity semantics, however, is only half of the problem, because different tasks depend on different semantic aspects.
Crime is tied to social and nightlife dynamics, whereas population is driven by commuting and residential activity, so a single task-agnostic embedding cannot serve all objectives equally well. 
Purpose-related semantic channels should therefore be constructed once and then selected per task, based on actual task needs.

Building on these observations, we propose EASTAR, an \underline{E}vent-\underline{A}nchored \underline{S}emantic and \underline{T}ask-\underline{A}daptive \underline{R}egion representation learning framework that 
constructs event-anchored, purpose-related semantic channels over aggregated mobility representations and routes them to each downstream task.
This is non-trivial: event signals are sparse and observed only at the region level, the aggregated flow entangles many purposes into one interaction strength, and the task-to-purpose mapping is unlabeled.

EASTAR proceeds in two decoupled stages. The first stage constructs the channels: it profiles each region's activities from offline events, propagates them along mobility links to offset sparse coverage, and 
uses these activity semantics as probes to factorize mobility representations into function-related semantic channels.
The representation also includes free channels for patterns events do not capture and a global channel for overall structure, with alignment and diversity constraints keeping the channels grounded in observed activity and mutually distinct. The second stage adapts the channels to each task: a large language model (LLM) turns the task description and channel semantics into a prior over channels, which initializes a lightweight router that is then refined under downstream supervision, so the prior sets only a starting direction while the final selection stays data-driven.

To summarize, this paper makes the following contributions:

\begin{itemize}
    \item We identify and characterize a semantic gap in mobility-based region representation: existing methods capture the structural dimension of inter-region interaction but largely overlook the activity-purpose semantics conflated within aggregated flow.

    \item We propose EASTAR, which treats offline events as region-level semantic anchors rather than an additional fusion view, using them to factorize aggregated mobility representations into event-anchored, purpose-related semantic channels.

    \item We design a task-adaptive routing mechanism that initializes task-channel dependencies from LLM-derived priors and refines them through downstream supervision, yielding interpretable, task-specific channel selection.

    \item We conduct extensive experiments on two real-world cities across four downstream tasks; EASTAR consistently outperforms representative baselines, and ablations confirm the value of event anchors, free channels, and task-adaptive routing.
\end{itemize}






\section{Related Work}

\subsection{Attribute-Oriented Region Representation Learning}



Attribute-oriented methods learn region representations from static urban attributes, mitigating the reliance on large-scale mobility data that is often restricted by privacy constraints. These approaches typically exploit public data sources, including POI distributions, land-use patterns, satellite imagery, street-view images, and urban infrastructure records.

HGI~\cite{hgi} constructs hierarchical embeddings at the POI, region, and city levels using graph neural networks, where region embeddings are aggregated from POI-level representations. CityFM~\cite{cityfm} models structural relations between urban entities extracted from OpenStreetMap and integrates multi-source contrastive signals for representation learning. RegionDCL~\cite{regiondcl} employs convolutional neural network (CNN)-based contrastive learning on building footprints to model fine-grained spatial structures and aggregates building-level semantics into region embeddings.

Tile2Vec~\cite{tile2vec} directly applies CNN to satellite imagery to learn spatial embeddings. FlexiReg~\cite{flexireg} learns grid-based region representations from heterogeneous urban attributes and introduces adaptive aggregation for downstream tasks. Other studies explore semi-supervised~\cite{SemiGTX} or generative~\cite{urbanverse} approaches for fine-grained or cross-city region representation. Recent studies also incorporate LLMs; for instance, UrbanLN~\cite{urbanln} aligns street-view images with LLM-generated semantic descriptions through vision-language pre-training.

Although these methods enrich region representations with static urban context, their semantic signals are primarily region-internal: they describe what facilities, physical structures, or visual patterns exist within a region. They do not explicitly model the activity-purpose semantics carried by inter-region mobility interactions.

\subsection{Mobility-Based Interaction Representation Learning}

Mobility-based methods learn urban region representations by modeling inter-region interaction relations, where human mobility serves as the primary observational signal for capturing functional dependencies. Representative methods typically construct region graphs from origin--destination (OD) flow matrices or mobility trajectories and learn embeddings through graph-based, contrastive, or multi-view fusion mechanisms. While some of these methods incorporate static attributes such as POIs and land use as auxiliary features, their core representational capacity is driven by inter-region mobility topology.

ZE-Mob~\cite{zemob} calculates pointwise mutual information by measuring regional co-occurrences in mobility trajectories and learns embeddings whose dot products approximate these values. HDGE~\cite{hdge} constructs mobility and spatial similarity graphs, then applies random walks and Skip-gram to learn region embeddings. MVURE~\cite{mvure} builds independent graphs from mobility, POI similarity, and check-in behavior, and employs graph attention networks (GATs) to learn view-specific embeddings before fusing them through weighted aggregation.

MGFN~\cite{mgfn} constructs multiple mobility graphs at different time steps, groups them into mobility patterns based on temporal similarity, and models intra- and inter-pattern correlations using self-attention mechanisms. HAFusion~\cite{hafusion} further improves multi-view fusion through hierarchical attention, where region-level and view-level correlations are explicitly encoded into the learned embeddings. ReCP~\cite{recp} enhances cross-view consistency via contrastive learning, while ComSRE~\cite{comsre} models mobility as shared and view-specific components to obtain disentangled representations. Recent work extends mobility-based interaction modeling to multimodal and cross-city settings, such as UrbanMMCL~\cite{urbanmmcl} and CRRL~\cite{crrl}.

These methods effectively capture mobility topology and latent interaction patterns, but the learned dimensions are usually not grounded in observable activity-purpose semantics. Even methods that disentangle mobility or multi-view signals mainly separate temporal patterns, views, or shared/private components, rather than constructing event-anchored semantic channels within aggregated mobility representations.

\subsection{Task-Adaptive Region Representation Learning}

Beyond representation learning itself, another line of research studies how to adapt urban region embeddings to specific downstream tasks. Since different tasks rely on different functional semantics, task-agnostic region embeddings may not be equally suitable for all prediction objectives. For example, crime prediction may be related to social and nightlife activity patterns, while population estimation may depend more on residential and commuting-related signals. This motivates task-adaptive region representation learning.
HREP~\cite{hrep} introduces prompt learning by attaching task-specific learnable prefixes to region embeddings, which are then fine-tuned for downstream tasks. GURPP~\cite{gurpp} integrates prompt mechanisms into graph-based urban representation learning to enhance task alignment. ToPT~\cite{topt} proposes a task-oriented prompt tuning framework that injects spatial priors via a Graphormer-based encoder, processes task descriptions using a frozen multimodal LLM (MLLM), and aligns task semantics with region embeddings through cross-attention. UrbanMoE~\cite{urbanmoe} constructs a multimodal multi-task framework that jointly models multiple urban prediction tasks using satellite imagery, POI features, and LLM-generated textual descriptions. It adopts a sparse Mixture-of-Experts architecture to dynamically select task-specific experts and reduce negative transfer across tasks.

These methods improve the flexibility of region representations through adaptive aggregation, prompt tuning, feature alignment, or expert selection mechanisms. The mapping between task requirements and underlying urban functional semantics remains largely implicit. 
EASTAR differs from these works by factorizing aggregated mobility representations into event-anchored semantic channels and using LLM-derived priors only to initialize task-channel routing, which is further refined by downstream supervision.


In summary, existing methods lack an explicit mechanism for grounding mobility-based region representations in activity-purpose semantics before task adaptation. EASTAR addresses this gap by using offline events as sparse semantic anchors to factorize mobility representations into interpretable channels for task-adaptive routing.

\section{Preliminaries}
In this section, we formalize the multi-source urban data used in this study and define the problem of task-adaptive urban region representation learning.

\textbf{Regions.} We partition a city into a set of $N$ non-overlapping spatial units based on standard administrative boundaries, denoted as $\mathcal{R}=\{r_1,r_2,\ldots,r_N\}$. Each region $r_i \in \mathcal{R}$ serves as the fundamental spatial entity for representation learning. We consider four types of features for each region as follows.

\textbf{POI features.} For each region $r_i$, we aggregate the frequency of POIs across distinct functional categories sourced from OpenStreetMap\cite{osm}. The POI feature is denoted as $\mathbf{x}_i^{poi} \in \mathbb{R}^{C_{poi}}$, where $C_{poi}$ is the number of POI categories.

\textbf{Land use features.} For each region $r_i$, we count the number of zones belonging to diverse land use types. The land use feature vector for region $r_i$ is formalized as $\mathbf{x}_i^{land} \in \mathbb{R}^{C_{land}}$, where $C_{land}$ represents the total number of land use types.

\textbf{Human mobility features.} We define a human mobility matrix $\mathbf{A}^{mob} \in \mathbb{R}^{N \times N}$ from aggregated taxi trip records, where $A^{mob}_{i,j}$ denotes the number of trips from region $r_i$ to region $r_j$ over an observation period. The $i$-th row of $\mathbf{A}^{mob}$ serves as the mobility feature of region $r_i$.

\textbf{Visual features.} To provide external environmental context for downstream task adaptation, we collect satellite and street-view imagery for each region $r_i$, denoted as $\mathcal{I}_i$.

\textbf{Event semantic features.} We introduce event data from offline activity platforms to capture the functional semantics of regional activities. Raw event categories are consolidated into $K = 8$ semantic anchors, as detailed in Table \ref{tab:semantic_aggregation}, grouping categories that share similar spatial distribution patterns while separating those with distinct trip motivations. For each region $r_i$, we define a semantic distribution $\mathbf{s}_i \in \mathbb{R}^K$ quantifying the normalized activity intensity across these $K$ anchors.


\begin{table}[htbp]
\centering
\footnotesize 
\renewcommand{\arraystretch}{1.0} 
\caption{Activity Categories Grouped into Semantic Anchors}
\vspace{-2 mm}
\label{tab:semantic_aggregation}
\begin{tabularx}{\columnwidth}{@{} c | >{\raggedright\arraybackslash}X @{}} 
\toprule 
\textbf{Semantic Anchor} & \textbf{Associated Activity Categories}
\\
\midrule 
\textbf{Tech $c_1$}      & Career/Business, Tech
\\ 
\textbf{Social $c_2$}    & Cars/Motorcycles, Dancing, Games, Singles, Socializing
\\ 
\textbf{Arts $c_3$}      & Fashion/Beauty, Fine Arts/Culture, Writing, Hobbies/Crafts, Movies/Film, Music, Photography, Sci-Fi/Fantasy
\\ 
\textbf{Health $c_4$}    & Health/Wellbeing, Outdoors/Adventure, Support, Fitness, Sports/Recreation
\\ 
\textbf{Education $c_5$} & Book Clubs, Education/Learning
\\ 
\textbf{Community $c_6$} & Movements/Politics, Parents/Family, Community/Environment, Pets/Animals
\\ 
\textbf{Food $c_7$}      & Food/Drink
\\ 
\textbf{Diversity $c_8$} & LGBT, Alternative Lifestyle, Language/Ethnic Identity, New Age/Spirituality, Paranormal, Religion/Beliefs \\
\bottomrule 
\end{tabularx}
\end{table}

\textbf{Problem Definition.} Given a set of non-overlapping urban regions $\mathcal{R} = \{r_1, \dots, r_N\}$ with multi-source feature matrices and offline event records, we aim to learn a region embedding function $f: (r_i, t) \rightarrow \mathbf{h}_i^{(t)} \in \mathbb{R}^d$ that maps each region to a compact vector adaptable to a downstream task $t$. The learned representations are expected to capture both structural interactions and functional semantics of regions, supporting diverse urban prediction tasks such as crime prediction.

\begin{figure*}[htbp]
    \centering
    \includegraphics[width=1\linewidth]{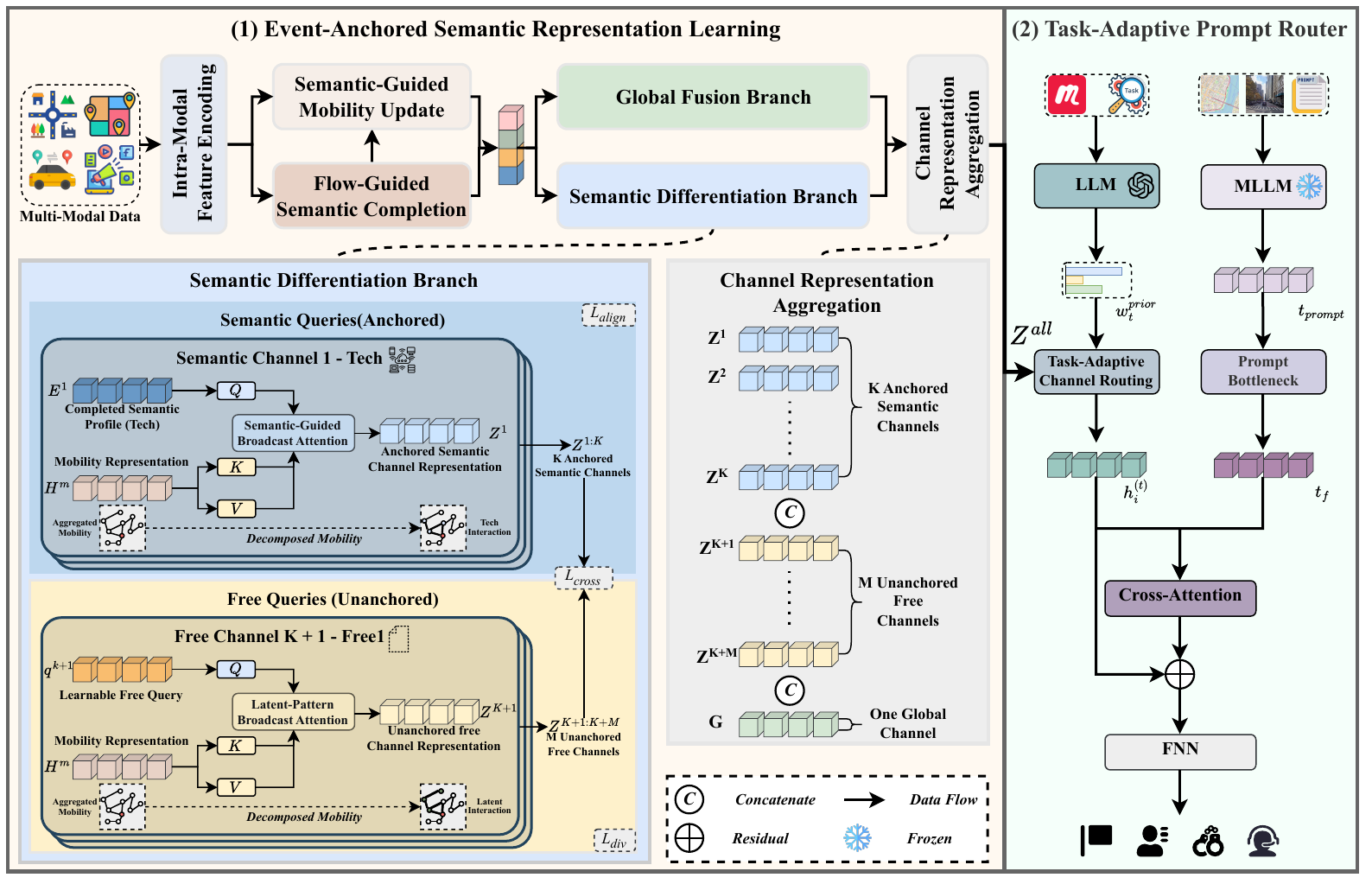}
    \vspace{-7 mm}
    \caption{Overview of EASTAR. Stage 1 learns event-anchored multi-channel region representations through semantic completion and channel decomposition, while Stage 2 adaptively routes the learned channels to downstream tasks.}
    \label{fig:framework}
    \vspace{-3 mm}
\end{figure*}

\section{Proposed Model}
In this section, we present our proposed model, EASTAR. We first outline the overall framework, then describe the two core stages: the Event-Anchored Semantic Representation Learning and the Task-Adaptive Prompt Router in detail. Finally, we introduce the decoupled training objectives.

\subsection{Model Overview}

Fig.~\ref{fig:framework} illustrates the overall architecture of EASTAR, which consists of two decoupled stages.

Stage 1: \underline{E}vent-\underline{A}nchored \underline{S}emantic \underline{R}epresentation \underline{L}earning (EASRL).
The EASRL module first completes sparse event semantics via mobility-guided propagation, 
then constructs $K{+}M{+}1$  event-anchored, free, and global channels through a dual-branch architecture,
producing a task-agnostic multi-channel foundation embedding.

Stage 2: \underline{T}ask-\underline{A}daptive \underline{P}rompt \underline{R}outer (TAPR).
With the EASRL module frozen, TAPR adapts the shared channels to each downstream task in three steps: LLM-guided prior initialization, learnable channel routing, and prompt-conditioned refinement via multimodal urban signals.





\subsection{Event-Anchored Semantic Representation Learning}
As shown in Fig.~\ref{fig:framework} (1), EASRL constructs $K+M+1$ task-agnostic channel representations per region, shared across downstream tasks.

\subsubsection{Heterogeneous Modality Encoding}

Before constructing semantic channels, we first encode the basic heterogeneous urban modalities into latent structural representations.

\textbf{Structural Modality Encoding.} 
The POI matrix $\mathbf{X}^{poi} \in \mathbb{R}^{N \times C_{poi}}$, 
land-use matrix $\mathbf{X}^{land} \in \mathbb{R}^{N \times C_{land}}$, 
and mobility matrix $\mathbf{X}^{mob} \equiv \mathbf{A}^{mob} \in \mathbb{R}^{N \times N}$ 
are first projected into a latent space through modality-specific multilayer perceptrons (MLPs). 
For each of these modalities, denoted by $v$, 
we obtain the encoded representation
$\mathbf{H}^{v}$ through a self-attention encoder:
\begin{equation}
\mathbf{H}^{v}=\mathrm{SelfAttn}\big(\sigma(\mathbf{X}^{v}\mathbf{W}^{v}+\mathbf{b}^{v})\big),
\end{equation}
where $\mathbf{W}^{v}$ and $\mathbf{b}^{v}$ are learnable
projection parameters, and $\sigma$ denotes the rectified linear unit activation.

\textbf{Event Semantic Profiling.} 
To obtain semantic signals associated with different activity purposes,
we organize regional offline events into
$K$ activity categories
$\mathcal{C}=\{c_1,\ldots,c_K\}$
(Table~\ref{tab:semantic_aggregation}).
For each region $r_i$ and category $c_k$,
we aggregate all associated event records
(e.g., titles, descriptions, and participant counts)
into a category-specific text document $T_{i,k}$.
A pre-trained sentence transformer~\cite{sentencebert}
encodes and projects each document into the model space:
\begin{equation}
\mathbf{E}^{k}_i =
\begin{cases}
f_\theta(T_{i,k})\, \mathbf{W}_e,
& \text{if } c_k \text{ is observed in region } r_i,\\[4pt]
\mathbf{0},
& \text{otherwise,}
\end{cases}
\end{equation}
where $W_e \in \mathbb{R}^{d_{\text{text}} \times d}$ is a learnable matrix. Zero-initializing absent categories preserves the sparsity of regional event distributions and provides missing-category indicators for subsequent semantic completion. The resulting profile set
$\{\mathbf{E}_i^1, \ldots, \mathbf{E}_i^K\}$
serves as semantic anchors for the downstream modules.

\subsubsection{Semantic-Structural Co-Enhancement} Due to the uneven spatial coverage of offline events, we propose a bidirectional semantic-structural co-enhancement mechanism that propagates semantics through mobility flows and injects semantic awareness back into mobility representations.

\textbf{Flow-Guided Semantic Completion.}
Many regions contain incomplete event observations. We complete missing semantic signals by propagating event tokens over the mobility graph, assuming that mobility-connected regions tend to share related functional semantics.



We perform semantic propagation on the mobility graph 
$\mathbf{A}^{mob}$ using a flow-aware graph attention mechanism. 
Let $\mathbf{E} \in \mathbb{R}^{N \times K \times d}$ denote the semantic tokens. 
For each channel $k$, query and key vectors are computed as 
$\mathbf{q}^{sem}_{i,k} = \mathbf{E}^k_i \mathbf{W}^c_q$ 
and 
$\mathbf{k}^{sem}_{j,k} = \mathbf{E}^k_j \mathbf{W}^c_k$, 
where 
$\mathbf{W}_q^c,\mathbf{W}_k^c\in\mathbb{R}^{d\times d}$. 
The flow-aware attention score between
regions $i$ and $j$ is defined as:
\begin{equation}
S_{i,j}^k = \frac{\mathbf{q}^{sem}_{i,k} (\mathbf{k}^{sem}_{j,k})^\top}{\sqrt{d}} + \beta \log\bigl(1 + A^{mob}_{i,j}\bigr),
\end{equation}
where the first term measures semantic similarity,
while the second term biases propagation toward
regions connected by stronger mobility flows.
The parameter $\beta$ controls the strength of this flow prior.
To prevent noisy semantic transfer,
propagation is restricted to flow-connected neighbors
and self-loops.
Specifically, disconnected region pairs are masked before
softmax normalization.
The propagated semantic token is then obtained through
attention-weighted aggregation:
\begin{equation}
\mathbf{E}^{p,k}_{i}
=
\sum_{j=1}^{N}
\mathrm{softmax}_{j}(S^{k}_{i,j})
\,\mathbf{E}^{k}_{j}.
\end{equation}

This operation propagates semantic evidence from
mobility-connected regions while preserving
channel-specific activity semantics.
Stacking all channels yields the propagated semantic profile
$\mathbf{E}^{p}_{i}$.
Semantic completion is performed at the region level. Let $\mathbf{m}_i \in \{0,1\}^K$ denote the observation mask indicating which activity categories are observed in region $r_i$.
If all $K$ categories are observed,
its original semantic profile is retained;
otherwise, the propagated profile is adopted.
\begin{equation}
\mathbf{E}^{c}_{i} =
    \begin{cases}
    \mathbf{E}_{i}, & \text{if } \sum_{k=1}^{K} m_{i,k}=K
    \\
    \mathbf{E}^{p}_{i}, & \text{otherwise}
    \end{cases}
\end{equation}
This region-level replacement is adopted because propagation also refines observed channels through neighborhood consensus.
The resulting $\mathbf{E}^{c}$ provides complete semantic
prototypes for all regions and serves as input to the
subsequent module.



\textbf{Semantic-Guided Mobility Update.} 
After semantic completion, we inject the completed semantic tokens $\mathbf{E}^{c}$ back into mobility representations through cross-attention. This step enriches mobility features with functional context while preserving the original mobility topology through residual fusion.



Specifically, mobility representations
$\mathbf{H}^{m}$
serve as queries,
while the completed semantic tokens
$\mathbf{E}^{c}$
provide keys and values to compute the semantic-enhanced mobility features:
\begin{equation}
\Delta \mathbf{H}^{m} = \mathrm{Softmax}\left(\frac{(\mathbf{H}^{m}\mathbf{W}^u_q)(\mathbf{E}^{c}\mathbf{W}^u_k)^\top}{\sqrt{d}}\right)(\mathbf{E}^{c}\mathbf{W}^u_v),
\end{equation}
where
$\mathbf{W}^{u}_{q}$,
$\mathbf{W}^{u}_{k}$,
and
$\mathbf{W}^{u}_{v}$
are learnable projection matrices.
The resulting
$\Delta \mathbf{H}^{m}$
encodes semantic context relevant to each region, thereby enriching mobility representations with functional signals.

To preserve the original mobility topology,
we adopt a residual fusion strategy:
\begin{equation}
\mathbf{H}^{m}_{u} = \gamma \mathbf{H}^{m} + (1 - \gamma) \Delta \mathbf{H}^{m},
\end{equation}
where $\gamma$ is a learnable balancing coefficient.

The resulting
$\mathbf{H}^{m}_{u}$
is used by the global fusion branch,
providing mobility representations enriched with functional semantics.
In contrast, the semantic differentiation branch operates on the original mobility representation
$\mathbf{H}^{m}$,
preserving raw interaction patterns for channel decomposition.
This design prevents semantic injection from homogenizing the functional channels while still benefiting the global branch.

\subsubsection{Dual-Branch Multi-Channel Encoding}
To jointly preserve global spatial structure and disentangle function-specific semantics, we employ a dual-branch architecture consisting of a global fusion branch and a semantic differentiation branch.

\textbf{Global Fusion Branch.} To construct a task-agnostic structural representation, we adopt the dual-feature attentive fusion module (DAFusion) from HAFusion~\cite{hafusion} to integrate heterogeneous urban views. DAFusion consists of two cascaded sub-modules: ViewFusion and RegionFusion. The input views correspond directly to the three encoded structural modalities: semantic-enhanced mobility representation $\mathbf{H}^m_u$, POI representation $\mathbf{H}^{p}$, and land-use representation $\mathbf{H}^{l}$. Semantic profiles are explicitly excluded from this branch to maintain a task-agnostic structural foundation; however, semantic information is still implicitly present through the updated mobility $\mathbf{H}^m_u$.

ViewFusion models interactions across different views and produces view-specific representations $\hat{\mathbf H}^{v}$. Adaptive fusion weights $\alpha^{v}$ are then learned to aggregate the views:
\begin{equation}
\tilde{\mathbf{G}} = \sum_{v \in \mathcal{V}} \alpha^v \hat{\mathbf{H}}^v.
\end{equation}

RegionFusion further applies multi-head self-attention to the fused representation $\tilde{\mathbf G}$
to capture higher-order spatial dependencies among regions. 

The resulting representation of region $r_i$ is denoted by $\mathbf g_i\in\mathbb R^d$, which serves as the global structural embedding.

\textbf{Semantic Differentiation Branch.}
While the global branch preserves holistic spatial structure, different activity purposes remain entangled within the aggregated mobility representation. To disentangle these latent interaction patterns, we use completed semantic profiles as functional probes to factorize mobility representations into semantic channels.

Specifically, the completed semantic profiles $\mathbf E^c$ provide semantic guidance, while the mobility representation $\mathbf H^m$ serves as the interaction space to be factorized. For each semantic category $c_k$,
Broadcast Attention selectively extracts the mobility patterns most relevant to that activity purpose, thereby producing a dedicated semantic channel. For category $c_k$, the semantic token $\mathbf E^c_{i,k}$ acts as the query, while $\mathbf H^m$ is broadcast as keys and values:
\begin{equation}
\mathbf{z}^k_i = \mathrm{BroadcastAttn}\big(\mathbf{E}^{c}_{i,k},\,\mathbf{H}^{m},\,\mathbf{H}^{m}\big),
\end{equation}
where $\mathrm{BroadcastAttn}(\cdot)$ uses the semantic token as a region-specific query, while all regional mobility representations serve as shared keys and values. This design allows each semantic channel to retrieve function-relevant interaction patterns from the global mobility topology.

As a result, each channel $\mathbf z_i^k$ captures a distinct mobility interaction pattern associated with a specific functional purpose. 


\textbf{Free Channels.}
Although the anchored semantic channels capture activity-specific mobility patterns, event platforms provide only a partial view of urban behavior. According to Hägerstrand's time-geography theory~\cite{hagerstraand1970what}, a substantial proportion of urban mobility is driven by routine and constraint-based activities, such as commuting and daily maintenance, which are often absent from event-centric data sources.

To account for these latent interaction patterns, we introduce $M$ free channels parameterized by learnable query vectors $\{\mathbf q^{K+1},\ldots,\mathbf q^{K+M}\}$. Each free query extracts latent mobility patterns through the same Broadcast Attention operation:
\begin{equation}
\mathbf{z}^{K+m}_i = \mathrm{BroadcastAttn}(\mathbf{q}^{K+m}, \mathbf{H}^m, \mathbf{H}^m).
\end{equation}


Unlike semantic channels, free channels are not associated with predefined activity categories. Instead, they are optimized to capture latent interaction patterns beyond those represented by the anchored semantic channels.

\subsubsection{Channel Representation Aggregation}
The final output of the EASRL module for region $i$, denoted as $\mathbf{Z}_{all, i}$, is a structured set of $K+M+1$ channel representations that concatenates the $K$ event-anchored channels, the $M$ free channels, and the single global consensus channel $\mathbf{g}_i$. 

For the Stage-1 reconstruction objectives, these channels are temporarily aggregated into a foundation embedding via learnable channel weights, whereas the full channel set $\mathbf{Z}_{all,i}$ is preserved and transferred to Stage 2.

\subsubsection{Multi-Channel Semantic Regularization}
To encourage semantic interpretability, channel diversity, and functional disentanglement, we introduce three regularization objectives for the anchored and free channels.

\textbf{Semantic Alignment ($\mathcal{L}_{align}$).} 
To ensure that anchored channels capture their intended activity semantics, we align channel activations with the observed activity distribution on fully covered regions. Let $\mathcal{O}=\{j:\sum_{k=1}^{K}m_{j,k}=K\}$ denote the set of regions with complete semantic observations. The channel activation distribution is computed as:
\begin{equation}
p^k_j = \frac{\exp(\|\mathbf{z}^k_j\|_2)}{\sum_{c=1}^{K}\exp(\|\mathbf{z}^c_j\|_2)},
\end{equation}
and aligned with the distribution $\mathbf s_j$ through Kullback--Leibler (KL) divergence:
\begin{equation}
\mathcal{L}_{align}=\frac{1}{|\mathcal{O}|}\sum_{j\in\mathcal{O}}\mathrm{KL}(\mathbf{s}_j\,\|\,\mathbf{p}_j).
\end{equation}

This constraint is applied only to the $K$ anchored channels, while free channels remain unconstrained.

\textbf{Free Channel Diversity ($\mathcal{L}_{div}$).} To encourage specialization among free channels and avoid redundant representations, we penalize pairwise cosine similarity among $\ell_2$-normalized channel representations ($\bar{\mathbf z}=\mathbf z/\|\mathbf z\|_2$):
\begin{equation}
\mathcal{L}_{div} = \frac{1}{N} \sum_{i=1}^{N} \sum_{m=1}^{M} \sum_{\substack{m'=1\\ m'\neq m}}^{M} \left( \cos(\bar{\mathbf{z}}_{i}^{K+m}, \bar{\mathbf{z}}_{i}^{K+m'}) \right)^2,
\end{equation}

\textbf{Cross-Channel Decorrelation ($\mathcal{L}_{cross}$).} To encourage free channels to model interaction patterns beyond the predefined semantic anchors, we minimize their similarity to all anchored channels:
\begin{equation}
\mathcal{L}_{cross}=\frac{1}{NMK}\sum_{i=1}^{N}\sum_{m=1}^{M}\sum_{k=1}^{K}
\cos(\bar{\mathbf{z}}^{K+m}_i,\bar{\mathbf{z}}^k_i),
\end{equation}

Together, these objectives preserve semantic interpretability, promote channel diversity, and improve functional disentanglement across the multi-channel representation space.

\subsection{Task-Adaptive Prompt Router}
As shown in Fig.~\ref{fig:framework} (2), TAPR adapts the shared channels to downstream tasks through three steps: LLM-guided prior initialization, learnable channel routing, and prompt-conditioned refinement.

\subsubsection{LLM-Guided Prior Initialization}

Different downstream tasks rely on distinct functional semantics. To provide an informative initialization for channel routing, we leverage an LLM to estimate the relevance of each channel to a target task.

Given a task description and the semantic definitions of all channels, the LLM outputs a prior importance vector $\mathbf{w}_t^{prior}\in\mathbb{R}^{K+M+1}$, where each element indicates the relevance of a channel to task $t$. Since free channels are not associated with predefined semantic categories, they typically receive lower prior weights, allowing their importance to be determined during subsequent data-driven optimization.


\subsubsection{Task-Adaptive Channel Routing}

To adapt the LLM prior to task-specific data distributions, we introduce a learnable channel routing mechanism.

For each task $t$, we define a learnable routing parameter $\mathbf w_t \in \mathbb R^{K+M+1}$, initialized from the normalized prior:

\begin{equation}
\mathbf{w}_t \leftarrow \log(\mathbf{w}_t^{prior} + \epsilon).
\end{equation}

During training, $\mathbf w_t$ is optimized end-to-end. The routing weights are obtained through a temperature-scaled softmax:
\begin{equation}
\boldsymbol{\alpha}_t = \mathrm{Softmax}(\mathbf{w}_t / \tau),
\end{equation}
which are used to aggregate the channel representations:
\begin{equation}
\mathbf{h}^{(t)}_i = \sum_{c=1}^{K+M+1} \alpha_t^c \, \mathbf{z}_i^c, \qquad \mathbf{z}_i^{K+M+1} \equiv \mathbf{g}_i.
\end{equation}

This design enables explicit and differentiable selection of task-relevant semantic channels, allowing shared representations to be flexibly adapted to diverse downstream objectives.

\subsubsection{Prompt-Conditioned Task Alignment}
Channel routing selects task-relevant semantic channels, but it does not explicitly incorporate external task context. We therefore introduce a lightweight prompt-conditioned refinement module to provide complementary task guidance.

Given a task-specific prompt template along with the visual features $\mathcal{I}_i$, a frozen MLLM generates a task-aware prompt embedding $\mathbf t_p$. The detailed prompt templates and MLLM configurations are provided in the supplementary material. After bottleneck compression, the resulting representation $\mathbf t_f$ is aligned with the routed embedding $\mathbf h_i^{(t)}$ through cross-attention:
\begin{equation}
    \mathbf{h}_r = \mathrm{CrossAttn}(\mathbf{h}_i^{(t)}, \mathbf{t}_f, \mathbf{t}_f),
\end{equation}
where the routed representation serves as the query, while the task embedding provides keys and values. The resulting representation $\mathbf h_r$ encodes task-specific contextual information retrieved from the prompt embedding. 

The aligned task context is fused with the routed representation through a residual prediction head:
\begin{equation}
\hat{y}^{(t)}_i = \mathbf{w}_{b}^\top \mathbf{h}^{(t)}_i
+ \mathrm{MLP}_{\delta}(\mathbf{h}^{(t)}_i \,\|\, \mathbf{h}_{r}),
\end{equation}
where $\mathbf w_b^\top \mathbf h_i^{(t)}$ provides the base prediction, and $\mathrm{MLP}_{\delta}$ learns a task-specific residual correction.

\subsection{Model Training}


EASTAR is trained under a two-stage decoupled optimization framework: 
Stage 1 learns semantically structured task-agnostic representations, 
while Stage 2 performs lightweight task-specific adaptation.

\subsubsection{Stage 1: Event-Anchored Semantic Representation Learning}

The first stage learns a semantically structured yet task-agnostic representation space through two complementary objectives: structural reconstruction and multi-channel semantic regularization. Structural reconstruction preserves global spatial dependencies, whereas semantic regularization enforces interpretable and disentangled channel representations.

\textbf{Mobility-Oriented Objective.} To preserve the global interaction topology, we reconstruct the mobility graph from the foundation embedding obtained by aggregating all channels. We reconstruct the mobility adjacency matrix $\mathbf{A}^{mob}$ from the aggregated foundation embedding $\mathbf{R}_i$ via source and destination projections $\mathbf{u}^{src}_i, \mathbf{u}^{dst}_i$. Here, $A^{mob}_{i,j}$ 
denotes the mean-normalized trip count, treated as an interaction weight rather than a probability.

Given a source region $i$, the model defines a transition probability distribution over all $N$ destination regions via:
\begin{equation}
p(j|i) = \frac{\exp({\mathbf{u}^{src}_i}^\top \mathbf{u}^{dst}_j)}
{\sum_{v=1}^{N} \exp({\mathbf{u}^{src}_i}^\top \mathbf{u}^{dst}_v)},
\end{equation}
and symmetrically $p'(i|j)$ for the reverse direction. The bidirectional loss is:
\begin{equation}
\mathcal{L}_{mob} = -\sum_{i,j} A^{mob}_{i,j}[\log(p(j|i)+\epsilon) + \log(p(i|j)+\epsilon)],
\end{equation}
where $\epsilon = 10^{-4}$. This objective encourages the aggregated representation to preserve region-to-region mobility interactions encoded in the original flow network.

\textbf{Attribute-Oriented Objective.}
In addition to mobility interactions, we preserve static urban attributes by reconstructing similarity structures derived from POI and land-use features. For view $v \in \{poi, land\}$, we enforce consistency between similarity matrices $\mathbf{S}^{v} \in \mathbb{R}^{N \times N}$ and the embedding space:
\begin{equation}
\mathcal{L}_{v}=\frac{1}{N^2}\sum_{i=1}^{N}\sum_{j=1}^{N}
\big(S^{v}_{i,j}-\cos(\mathbf{u}^v_i,\mathbf{u}^v_j)\big)^2.
\end{equation}

The overall reconstruction objective combines mobility and attribute preservation:
\begin{equation}
\mathcal{L}_{recon} = \mathcal{L}_{mob} + \mathcal{L}_{poi} + \mathcal{L}_{land}.
\end{equation}

The final Stage-1 objective jointly optimizes structural reconstruction and semantic regularization:
\begin{equation}
\mathcal{L}_{stage1} = \mathcal{L}_{recon} + \lambda_1\mathcal{L}_{align} + \lambda_2\mathcal{L}_{div} + \lambda_3\mathcal{L}_{cross},
\end{equation}
where $\lambda_1, \lambda_2, \lambda_3$ balance semantic alignment and channel disentanglement.

\subsubsection{Stage 2: Task-Adaptive Prompt Routing}
In Stage 2, the EASRL module is frozen and only task-specific components are optimized, including the routing parameters, prompt-alignment module, and prediction head. Given the task-specific prediction $\hat{y}^{(t)}_i$, we adopt the mean squared error loss:
\begin{equation}
\mathcal{L}_{stage2} = \frac{1}{|\mathcal{D}_{train}|} \sum_{i \in \mathcal{D}_{train}}
\left( \hat{y}^{(t)}_i - y_i \right)^2,
\end{equation}
where $y_i$ is the ground-truth label for region $i$ and $\mathcal{D}_{train}$ denotes the training set.


\section{Experiments}
We perform an empirical study to verify: 
(\textbf{Q1}) the prediction accuracy of our EASTAR as compared with the state-of-the-art (SOTA) methods on four downstream tasks, 
(\textbf{Q2}) the effectiveness of our core architectural components,
(\textbf{Q3}) the necessity of explicit channel decomposition and the impact of semantic anchor sources,
(\textbf{Q4}) the task-adaptive activation of our routing mechanism and the region-level consistency of the learned channels, and 
(\textbf{Q5}) the robustness across different LLMs and the sensitivity to key hyper-parameters.

\begin{table}[h]
\caption{Dataset Statistics}
\label{tab:dataset_statistics}
\vspace{-2 mm}
\centering
\small
\resizebox{\columnwidth}{!}{
\begin{tabular}{l|rr|r}
\toprule
\textbf{Dataset} & \textbf{NYC} & \textbf{CHI} & \textbf{Source} \\
\midrule
Regions & 180 & 77 & Open Portal~\cite{nyc_data,chicago_data}\\ 
POIs & 24,496 & 57,891 & OpenStreetMap~\cite{osm}\\
POI categories & 26 & 26 & OpenStreetMap~\cite{osm}\\
Land-use categories & 11 & 12 &  OpenStreetMap~\cite{osm}\\
Taxi trips & 10,953,879 & 3,381,807 & Open Portal~\cite{nyc_data,chicago_data}\\
Satellite images & 180 & 77 &  Google Maps~\cite{googlemaps}\\
Street view photos & 11,629 & 5,005 & Google Maps~\cite{googlemaps}\\
Meetup events& 147,327 & 45,546 & Meetup~\cite{meetup}\\
\midrule
Crime records & 35,335 & 18,200 &  Open Portal~\cite{nyc_data,chicago_data}\\
Check-in counts & 106,902 & 167,232 &Foursquare~\cite{foursquare}\\
Service call records & 516,187 & 24,350 & Open Portal~\cite{nyc_data,chicago_data}\\ 
Population counts & 1,540,692 & 2,508,984 & WorldPop~\cite{worldpop}\\
\bottomrule
\end{tabular}
} 
\end{table}

\subsection{Experimental Settings}
\textbf{Datasets.}
We conduct experiments on two real-world urban datasets: New York City (\textbf{NYC})~\cite{nyc_data} and Chicago (\textbf{CHI})~\cite{chicago_data}, spatially partitioned into 180 and 77 disjoint regions, respectively. Table~\ref{tab:dataset_statistics} summarizes the dataset statistics. For each city, we collect four types of regional features:
(1)~\textit{Human Mobility}: origin-destination flow matrices constructed from taxi trip records~\cite{nyc_data, chicago_data}, where each entry denotes the number of trips between a region pair over the observation period;
(2)~\textit{POI Distributions}: category-wise counts of Points-of-Interest extracted from OpenStreetMap~\cite{osm}, with 26 functional categories;
(3)~\textit{Land Use}: land-use category distributions sourced from OpenStreetMap~\cite{osm}, with 11 and 12 categories for NYC and CHI, respectively;
(4)~\textit{Social Events}: offline event records from the Meetup platform~\cite{meetup}, including activity categories, timestamps, and participant counts. The original 33 categories are consolidated into 8 semantic anchors.

We evaluate on four standard downstream tasks: predicting region-level counts of \textit{Crime}~\cite{nyc_data, chicago_data}, \textit{Check-ins}~\cite{foursquare}, \textit{Service Calls}~\cite{nyc_data, chicago_data}, and \textit{Population}~\cite{worldpop}. Following prior works~\cite{hafusion, flexireg, hrep}, we use these counts as regression targets.

\textbf{Evaluation Metrics.}
Following prior works~\cite{topt, hrep, flexireg}, all results use ten-fold cross-validation. EASTAR and other label-dependent task-adaptive methods are trained independently within each fold using only training-fold labels; Ridge only initializes EASTAR's base predictor. Representation-only baselines use fold-specific Ridge regression with the same splits. We report mean absolute error (\textbf{MAE}), root mean squared error (\textbf{RMSE}), and coefficient of determination $\mathbf{R^2}$.

\textbf{Baselines.} We compare EASTAR against 12 representative methods spanning three categories. 
For all baselines, we adopt the hyperparameter settings reported in the original papers.

\textbf{(1) Attribute-oriented methods:}

\begin{itemize}
    \item \textbf{UrbanCLIP}~\cite{urbanclip} applies contrastive learning on satellite image-text pairs to produce text-enhanced visual region embeddings.
    
    \item \textbf{RegionDCL}~\cite{regiondcl} learns region embeddings via intra-group and group-region contrastive learning over building footprints.
    
    \item \textbf{CityFM}~\cite{cityfm} learns entity embeddings from OpenStreetMap via contrastive objectives and aggregates them into region embeddings.

    \item \textbf{FlexiReg}~\cite{flexireg} (SOTA) learns grid-cell embeddings from multi-modal urban data with adaptive aggregation and prompt-based task adaptation.
\end{itemize}

\textbf{(2) Interaction-oriented methods:}
\begin{itemize}
    \item \textbf{MVURE}~\cite{mvure} applies GAT on mobility, POI, and check-in graphs and fuses view-specific embeddings via weighted summation.

    \item \textbf{MGFN}~\cite{mgfn} clusters multiple mobility graphs into pattern groups and applies message passing to generate region embeddings.

    \item \textbf{ReCP}~\cite{recp} integrates mobility and POI views by maximizing inter-view mutual information and minimizing conditional entropy.

    \item \textbf{HAFusion}~\cite{hafusion} fuses mobility, POI, and land-use views via hierarchical attention at both region and view levels.

    \item \textbf{ComSRE}~\cite{comsre} disentangles multi-view data into shared and view-specific components via contrastive alignment and orthogonality constraints.
\end{itemize}

\textbf{(3) Task-adaptive methods:}
\begin{itemize}
    \item \textbf{HREP}~\cite{hrep} prepends learnable prompt embeddings to region representations for task-specific fine-tuning.

    \item \textbf{GURPP}~\cite{gurpp} integrates prompt-based learning with graph-structured urban representations for task alignment.
    
    \item \textbf{ToPT}~\cite{topt} (SOTA) aligns Graphormer-based region embeddings with MLLM-generated task prompts via cross-attention.
    
\end{itemize}

\begin{table*}[htbp]
\caption{Performance comparison on four downstream tasks. (Best: \textbf{Bold}, Second: \underline{Underline}. $^*$ indicates a statistically significant improvement over the best baseline ($p < 0.05$). The `Improvement' is calculated against the best baseline.)}
\vspace{-2 mm}
\label{tab:overall_results}
\centering
\footnotesize
\setlength{\tabcolsep}{3pt}
\resizebox{\textwidth}{!}{
\begin{tabular}{l | ccc | ccc | ccc | ccc}
\toprule[1.2pt] 
\multirow{2}{*}{\textbf{New York City}} 
& \multicolumn{3}{c|}{Crime} 
& \multicolumn{3}{c|}{Check-in} 
& \multicolumn{3}{c|}{Service Call} 
& \multicolumn{3}{c}{Population} \\
\cmidrule{2-13}
& MAE $\downarrow$ & RMSE $\downarrow$ & $R^2\uparrow$ 
& MAE $\downarrow$ & RMSE $\downarrow$ & $R^2\uparrow$ 
& MAE $\downarrow$ & RMSE $\downarrow$ & $R^2\uparrow$ 
& MAE $\downarrow$ & RMSE $\downarrow$ & $R^2\uparrow$ \\
\midrule
MVURE\cite{mvure}             & 67.9 $\pm$ 1.1 & \phantom{0}93.8 $\pm$ 1.9 & 0.591 $\pm$ 0.016 & 306.7 $\pm$ \phantom{0}8.2 & 499.6 $\pm$ 12.9 & 0.627 $\pm$ 0.019 & 1428 $\pm$ 33 & 2180 $\pm$ \phantom{0}46 & 0.367 $\pm$ 0.027 & 2814 $\pm$ 67 & 3564 $\pm$ \phantom{0}79 & 0.545 $\pm$ 0.008 \\
MGFN\cite{mgfn}              & 70.2 $\pm$ 2.3 & \phantom{0}89.6 $\pm$ 2.5 & 0.630 $\pm$ 0.020 & 292.6 $\pm$ 17.1 & 451.8 $\pm$ 28.1 & 0.690 $\pm$ 0.040 & 1554 $\pm$ 81 & 2286 $\pm$ 115 & 0.303 $\pm$ 0.069 & 2822 $\pm$ 42 & 3706 $\pm$ \phantom{0}98 & 0.509 $\pm$ 0.026 \\
ReCP\cite{recp}              & 83.1 $\pm$ 2.4 & 108.7 $\pm$ 2.2 & 0.459 $\pm$ 0.022 & 246.9 $\pm$ \phantom{0}4.3 & 400.1 $\pm$ 23.6 & 0.761 $\pm$ 0.029 & 1516 $\pm$ 66 & 2199 $\pm$ \phantom{0}49 & 0.356 $\pm$ 0.029 & 3305 $\pm$ 90 & 4353 $\pm$ 153 & 0.322 $\pm$ 0.047 \\
HAFusion\cite{hafusion}          & 56.1 $\pm$ 1.3 & \phantom{0}76.1 $\pm$ 2.2 & 0.734 $\pm$ 0.015 & 202.8 $\pm$ \phantom{0}7.2 & 322.8 $\pm$ 12.6 & 0.844 $\pm$ 0.012 & 1273 $\pm$ 20 & 1951 $\pm$ \phantom{0}27 & 0.493 $\pm$ 0.014 & 2497 $\pm$ 50 & 3277 $\pm$ \phantom{0}82 & 0.616 $\pm$ 0.019 \\
ComSRE\cite{comsre}           & 72.3 $\pm$ 1.1 & \phantom{0}99.6 $\pm$ 6.5 & 0.545 $\pm$ 0.059 & 251.7 $\pm$ 17.1 & 434.8 $\pm$ 44.4 & 0.716 $\pm$ 0.057 & 1459 $\pm$ 69 & 2254 $\pm$ \phantom{0}98 & 0.324 $\pm$ 0.058 & 2964 $\pm$ 95 & 3916 $\pm$ 100 & 0.452 $\pm$ 0.028 \\
\midrule
UrbanCLIP\cite{urbanclip}         & 97.4 $\pm$ 2.6 & 126.1 $\pm$ 1.9 & 0.267 $\pm$ 0.012  & 393.6 $\pm$ \phantom{0}5.9  & 602.4 $\pm$ \phantom{0}3.1  & 0.458 $\pm$ 0.005  & 1409 $\pm$ \phantom{0}7   & 2401 $\pm$ \phantom{0}15   & 0.232 $\pm$ 0.005  & 3338 $\pm$ 11   & 4499 $\pm$ \phantom{0}16   & 0.276 $\pm$ 0.002 \\
RegionDCL\cite{regiondcl}         & 98.7 $\pm$ 0.7 & 127.9 $\pm$ 1.1 & 0.251 $\pm$ 0.025  & 371.2 $\pm$ 10.3  & 595.5 $\pm$ \phantom{0}5.9  & 0.471 $\pm$ 0.023  & 1783 $\pm$ 22   & 2597 $\pm$ \phantom{0}18   & 0.103 $\pm$ 0.026  & 3753 $\pm$ 47   & 4734 $\pm$ \phantom{0}59   & 0.198 $\pm$ 0.019 \\
CityFM\cite{cityfm}             
& 95.5 $\pm$ 1.4 & 122.4 $\pm$ 1.8 & 0.315 $\pm$ 0.010  
& 380.2 $\pm$ \phantom{0}3.8  & 594.9 $\pm$ \phantom{0}6.4  & 0.471 $\pm$ 0.011  
& 1780 $\pm$ 27   & 2577 $\pm$ \phantom{0}19   & 0.117 $\pm$ 0.013  
& 3515 $\pm$ 18   & 4545 $\pm$ \phantom{0}26   & 0.261 $\pm$ 0.002 \\
FlexiReg\cite{flexireg}          
& \underline{53.2 $\pm$ 0.4} & \underline{\phantom{0}73.6 $\pm$ 1.6} & \underline{0.751 $\pm$ 0.009} 
& 198.2 $\pm$ \phantom{0}3.6 & 309.0 $\pm$ 17.2 & 0.850 $\pm$ 0.011 
& \underline{1303 $\pm$ 16} & 1989 $\pm$ \phantom{0}43 & \underline{0.497 $\pm$ 0.012}
& \underline{2231 $\pm$ 24} & \underline{2974 $\pm$ \phantom{0}60} & \underline{0.701 $\pm$ 0.002} \\
\midrule
HREP\cite{hrep}              & 62.8 $\pm$ 2.1 & \phantom{0}83.1 $\pm$ 2.3  & 0.681 $\pm$ 0.014  & 276.3 $\pm$ 11.7  & 448.2 $\pm$ 16.9  & 0.700 $\pm$ 0.022  & 1430 $\pm$ 29   & 2127 $\pm$ \phantom{0}33   & 0.398 $\pm$ 0.019  & 2656 $\pm$ 59   & 3461 $\pm$ \phantom{0}83   & 0.571 $\pm$ 0.021 \\
GURPP\cite{gurpp}
& 72.9 $\pm$ 2.3    & \phantom{0}94.7 $\pm$ 4.6  & 0.589 $\pm$ 0.039  
& 267.0 $\pm$ 15.2  & 403.3 $\pm$ 16.5           & 0.757 $\pm$ 0.019  
& 1480 $\pm$ 33     & 2115 $\pm$ \phantom{0}84   & 0.405 $\pm$ 0.048  
& 3316 $\pm$ 75     & 4159 $\pm$ \phantom{0}87   & 0.381 $\pm$ 0.026 
\\
ToPT\cite{topt}              
& 54.7 $\pm$ 3.9  & \phantom{0}74.9 $\pm$ 5.6 & 0.742 $\pm$ 0.038    
& \underline{178.7 $\pm$ 12.5}& \underline{260.5 $\pm$ 16.0} & \underline{0.899 $\pm$ 0.012}   
& 1315 $\pm$ 29 & \underline{1956 $\pm$ 106} & 0.491 $\pm$ 0.056   
& 2293 $\pm$ 58 & 3049 $\pm$ \phantom{0}58  & 0.668 $\pm$ 0.012
\\
\midrule
\textbf{EASTAR$^*$}
& \textbf{45.7 $\pm$ 0.4} & \textbf{\phantom{0}57.8 $\pm$ 1.5} & \textbf{0.847 $\pm$ 0.008} 
& \textbf{137.4 $\pm$ \phantom{0}2.7} & \textbf{185.4 $\pm$ \phantom{0}8.7} & \textbf{0.949 $\pm$ 0.005} 
& \textbf{1107 $\pm$ 13} & \textbf{1604 $\pm$ \phantom{0}20} & \textbf{0.658 $\pm$ 0.015} 
& \textbf{2036 $\pm$ 44} & \textbf{2676 $\pm$ \phantom{0}81} & \textbf{0.744 $\pm$ 0.016} \\
\midrule
\textbf{Improvement}
& \textbf{14.10\%} & \textbf{21.47\%} & \textbf{12.78\%} 
& \textbf{23.11\%} & \textbf{28.83\%} & \textbf{5.56\%} 
& \textbf{13.04\%} & \textbf{17.79\%} & \textbf{32.39\%} 
& \textbf{8.74\%} & \textbf{10.02\%} & \textbf{6.13\%} \\
\bottomrule[0.75pt] 
\addlinespace[4pt]  
\toprule[1.2pt] 
\multirow{2}{*}{\textbf{Chicago}}
& \multicolumn{3}{c|}{Crime} 
& \multicolumn{3}{c|}{Check-in} 
& \multicolumn{3}{c|}{Service Call} 
& \multicolumn{3}{c}{Population} \\
\cmidrule{2-13}
& MAE $\downarrow$ & RMSE $\downarrow$ & $R^2\uparrow$ 
& MAE $\downarrow$ & RMSE $\downarrow$ & $R^2\uparrow$ 
& MAE $\downarrow$ & RMSE $\downarrow$ & $R^2\uparrow$ 
& MAE $\downarrow$ & RMSE $\downarrow$ & $R^2\uparrow$ \\
\midrule
MVURE\cite{mvure} & 100.4 $\pm$ 6.6\phantom{0} & 129.2 $\pm$ 7.3\phantom{0} & 0.461 $\pm$ 0.062 & 1693 $\pm$ \phantom{0}74 & 3171 $\pm$ 128 & 0.656 $\pm$ 0.029 & 190.3 $\pm$ \phantom{0}9.8 & 266.9 $\pm$ 12.1 & 0.441 $\pm$ 0.050 & 13717 $\pm$ 322 & 17174 $\pm$ 552 & 0.313 $\pm$ 0.043 \\
MGFN\cite{mgfn} & 107.4 $\pm$ 5.4\phantom{0} & 137.9 $\pm$ 5.2\phantom{0} & 0.386 $\pm$ 0.047 & 1281 $\pm$ \phantom{0}41 & 2276 $\pm$ \phantom{0}86 & 0.817 $\pm$ 0.011 & 208.2 $\pm$ 11.3 & 293.4 $\pm$ 16.6 & 0.329 $\pm$ 0.077 & 13071 $\pm$ 505 & 16578 $\pm$ 707 & 0.359 $\pm$ 0.054 \\
ReCP\cite{recp} & \phantom{0}86.9 $\pm$ 5.5\phantom{0} & 120.1 $\pm$ 7.1\phantom{0} & 0.534 $\pm$ 0.057 & 1272 $\pm$ \phantom{0}92 & 2341 $\pm$ 267 & 0.804 $\pm$ 0.045 & 206.7 $\pm$ 11.1 & 303.4 $\pm$ 16.1 & 0.284 $\pm$ 0.076 & 12085 $\pm$ 400 & 17029 $\pm$ 561 & 0.325 $\pm$ 0.044 \\
HAFusion\cite{hafusion} & \phantom{0}77.8 $\pm$ 3.6\phantom{0} & 107.1 $\pm$ 5.4\phantom{0} & 0.631 $\pm$ 0.036 & \phantom{0}929 $\pm$ \phantom{0}62 & 1947 $\pm$ \phantom{0}75 & 0.870 $\pm$ 0.010 & 159.3 $\pm$ 13.9 & 222.0 $\pm$ 18.9 & 0.613 $\pm$ 0.067 & 10678 $\pm$ 390 & 13988 $\pm$ 548 & 0.544 $\pm$ 0.035 \\
ComSRE\cite{comsre} & \phantom{0}75.2 $\pm$ 4.2\phantom{0} & 100.3 $\pm$ 6.2\phantom{0} & 0.675 $\pm$ 0.040 & 1609 $\pm$ 286 & 3927 $\pm$ 327 & 0.453 $\pm$ 0.088 & 213.0 $\pm$ 11.7 & 286.2 $\pm$ 18.8 & 0.362 $\pm$ 0.085 & 12063 $\pm$ 688 & 15308 $\pm$ 841 & 0.475 $\pm$ 0.057 \\
\midrule
UrbanCLIP\cite{urbanclip}         & 101.6 $\pm$ 0.6\phantom{0} & 134.7 $\pm$ 1.7\phantom{0} & 0.416 $\pm$ 0.006  & 2611 $\pm$ \phantom{0}29   & 4884 $\pm$ \phantom{0}72   & 0.186 $\pm$ 0.024  & 183.2 $\pm$ \phantom{0}0.9  & 256.3 $\pm$ \phantom{0}1.8  & 0.491 $\pm$ 0.003  & 13328 $\pm$ \phantom{0}69  & 17498 $\pm$ \phantom{0}74  & 0.288 $\pm$ 0.006 \\
RegionDCL\cite{regiondcl}         & 121.7 $\pm$ 2.3\phantom{0} & 159.6 $\pm$ 2.8\phantom{0} & 0.179 $\pm$ 0.028  & 2427 $\pm$ 123   & 4184 $\pm$ 136   & 0.402 $\pm$ 0.042  & 195.7 $\pm$ \phantom{0}7.6  & 267.1 $\pm$ 10.1  & 0.445 $\pm$ 0.041  & 14289 $\pm$ 343  & 18653 $\pm$ 368  & 0.190 $\pm$ 0.032 \\
CityFM\cite{cityfm}             & 121.6 $\pm$ 1.8\phantom{0} & 157.1 $\pm$ 2.8\phantom{0} & 0.205 $\pm$ 0.018  & 1980 $\pm$ \phantom{0}64   & 3362 $\pm$ 109   & 0.614 $\pm$ 0.025  & 198.3 $\pm$ \phantom{0}3.7  & 280.1 $\pm$ \phantom{0}6.1  & 0.391 $\pm$ 0.027  & 13904 $\pm$ \phantom{0}37  & 17704 $\pm$ \phantom{0}56  & 0.271 $\pm$ 0.004 \\
FlexiReg\cite{flexireg}          
& \underline{\phantom{0}61.7 $\pm$ 3.5\phantom{0}}  & \underline{\phantom{0}85.1 $\pm$ 4.2\phantom{0}}  & \underline{0.766 $\pm$ 0.022}  
& \underline{\phantom{0}922 $\pm$ \phantom{0}76}    & 1775 $\pm$ 198   & 0.891 $\pm$ 0.024  
& \underline{121.1 $\pm$ \phantom{0}7.4}  & 178.2 $\pm$ \phantom{0}9.5  & 0.753 $\pm$ 0.026  
& \underline{\phantom{0}8126 $\pm$ 224}   & \underline{11395 $\pm$ 255}  & \underline{0.698 $\pm$ 0.014} \\
\midrule
HREP\cite{hrep}              
& \phantom{0}88.3 $\pm$ 6.4\phantom{0}  &114.4 $\pm$ 5.5\phantom{0} & 0.578 $\pm$ 0.041  
& 1702 $\pm$ \phantom{0}79   & 3300 $\pm$ 100   & 0.628 $\pm$ 0.023  
& 185.7 $\pm$ \phantom{0}6.1  & 262.2 $\pm$ \phantom{0}5.8  & 0.468 $\pm$ 0.021  
& 12063 $\pm$ 539  & 15397 $\pm$ 832  & 0.447 $\pm$ 0.061 \\
GURPP\cite{gurpp}             
& \phantom{0}94.0 $\pm$ 1.7\phantom{0}  & 122.1 $\pm$ 2.9\phantom{0} & 0.557 $\pm$ 0.021  
& 2473 $\pm$ 124   & 4165 $\pm$ 215   & 0.388 $\pm$ 0.054  
& 197.1 $\pm$ \phantom{0}4.3  & 270.2 $\pm$ \phantom{0}5.7  & 0.433 $\pm$ 0.025  
& 10983 $\pm$ 179  & 14513 $\pm$ 215  & 0.510 $\pm$ 0.031 \\
ToPT\cite{topt}              
& 70.7 $\pm$ 3.9 & 94.7 $\pm$ 3.5   & 0.711 $\pm$ 0.022 
& \phantom{0}956 $\pm$ \phantom{0}96& \underline{1615 $\pm$ 201} & \underline{0.910 $\pm$ 0.022}  
& 131.1  $\pm$ \phantom{0}9.7  & \underline{176.5 $\pm$ \phantom{0}1.9}& \underline{0.758 $\pm$ 0.005} 
& \phantom{0}9237 $\pm$ 652 & 12265 $\pm$ 993   &  0.662 $\pm$ 0.058\\
\midrule
\textbf{EASTAR$^*$}     
& \textbf{49.5 $\pm$ 3.0} & \textbf{64.9 $\pm$ 1.8} & \textbf{0.864 $\pm$ 0.029} 
& \textbf{\phantom{0}561 $\pm$ \phantom{0}33} & \textbf{\phantom{0}985 $\pm$ \phantom{0}96} & \textbf{0.967 $\pm$ 0.021} 
& \textbf{\phantom{0}98.6 $\pm$ \phantom{0}3.1}  & \textbf{135.6 $\pm$ \phantom{0}7.1} & \textbf{0.857 $\pm$ 0.016} 
& \textbf{\phantom{0}7315 $\pm$ \phantom{0}334} & \textbf{\phantom{0}9030 $\pm$ 392} & \textbf{0.818 $\pm$ 0.017} 
\\
\midrule
\textbf{Improvement}
& \textbf{19.77\%} & \textbf{23.74\%} & \textbf{12.79\%} 
& \textbf{39.15\%} & \textbf{39.01\%} & \textbf{6.26\%} 
& \textbf{18.58\%} & \textbf{23.17\%} & \textbf{13.06\%} 
& \textbf{9.98\%} & \textbf{20.76\%} & \textbf{17.19\%} 
\\
\bottomrule[0.75pt] 
\end{tabular}
}
\vspace{-5 mm}
\end{table*}
\textbf{Model hyperparameter settings.} Our model is trained using a two-stage approach, with each stage optimized for 3{,}000 epochs in full batches. During the first stage, the embedding dimension is set to $d = 256$. We set the numbers of anchored semantic and free channels to $K = 8$ and $M = 2$, respectively. We optimize the network using the Adam optimizer with a learning rate of $3\times10^{-4}$ and weight decay of $10^{-4}$ (NYC) and $10^{-3}$ (CHI). In the second stage, the channel fusion weights are initialized using task priors generated by GPT-5.4, and the prompt embeddings are extracted from Llama-3.2-11B-Vision-Instruct~\cite{llama3}. These are followed by data-driven fine-tuning on each downstream task with task-specific learning rates.

\subsection{Overall Results (Q1)}
The overall prediction results are reported in Table~\ref{tab:overall_results}. As confirmed by t-tests ($p < 0.05$) against the strongest baseline on all tasks, EASTAR demonstrates statistically significant improvements. We make the following observations.

(1) EASTAR consistently achieves the best performance across all tasks and cities. EASTAR outperforms all baselines on both NYC and CHI across all four downstream tasks, improving $R^2$ by up to 17.19\% on CHI Population. The improvements are particularly pronounced on behavior-intensive tasks such as Check-in prediction, where EASTAR achieves 23.11\% and 39.15\% MAE reductions on NYC and CHI compared to the strongest baselines. This is attributed to two design choices: (i) the semantic differentiation branch structures aggregated mobility representations into function-specific channels, 
distinguishing trips driven by different activity purposes; (ii) the task-adaptive prompt router selectively activates task-relevant channels, enabling precise alignment between semantic dimensions and downstream objectives.

(2) Task-adaptive and flexible baselines are competitive but remain limited. Methods with task adaptation (ToPT, HREP, GURPP) or flexible aggregation (FlexiReg) generally outperform conventional approaches. ToPT achieves second-best performance on Check-in across both cities by leveraging MLLM-based task prompts, and FlexiReg remains highly competitive on Crime and Population due to its fine-grained grid-level modeling. However, these methods are outperformed by EASTAR. Their adaptation mechanisms are either weakly structured (e.g., randomly initialized prompts in HREP) or operate on undifferentiated representations without explicit semantic decomposition, limiting task-specific alignment.

(3) Conventional interaction-oriented and attribute-oriented methods show clear gaps. Interaction-oriented methods (MVURE, MGFN, ReCP, HAFusion) model mobility as undifferentiated flow matrices, and even ComSRE, which disentangles shared and view-specific components, separates signals across views rather than decomposing the flow by activity purpose. Attribute-oriented methods (UrbanCLIP, RegionDCL, CityFM) rely on static features that cannot capture dynamic inter-region activity.


(4) EASTAR's advantage is more pronounced on CHI and on behavior-intensive tasks. The average MAE reduction on CHI (21.9\%) consistently exceeds that on NYC (14.7\%). 
We attribute this to CHI having fewer, larger regions, whose aggregated mobility mixes more heterogeneous trip purposes, so semantic decomposition yields larger marginal gains.
Across tasks, behavior-intensive predictions (Check-in, Service Call) gain more than spatially structured ones (Crime, Population); we analyze the mechanism behind this in Section V-D.


\textbf{Computational cost and multi-task reuse.} Since EASTAR introduces semantic completion, multi-channel encoding, and task-specific routing, we examine whether these components incur prohibitive computational overhead. Table IV reports the representation learning cost and the per-task adaptation cost on NYC and CHI. 
All time measurements are reported in seconds. For the downstream tasks, the reported running times include both the task-specific model training (e.g., prompt learning) and inference, with the specific inference times provided in parentheses. Note that the downstream tasks share identical input and output sizes for a fair comparison. 
EASTAR requires moderate representation learning time. It is slower than lightweight interaction-oriented models because of semantic completion and multi-channel encoding, but substantially faster than methods that rely on large-scale visual or foundation-model pretraining like UrbanCLIP. More importantly, the Stage-1 representation is learned once and reused across downstream tasks. During Stage 2, EASTAR optimizes only lightweight routing and prediction components, leading to lower per-task adaptation cost than prompt-heavy task-adaptive baselines such as ToPT and FlexiReg. These results indicate that EASTAR’s performance gains do not come with prohibitive computational overhead and that the framework is practical when multiple downstream urban tasks share the same learned representation.

\begin{table}[htbp]
\caption{Embedding Learning and Testing Times (seconds)}
\vspace{-2 mm}
\label{tab:running_time}
\centering
\small
\renewcommand{\arraystretch}{1} 
\resizebox{\columnwidth}{!}{
\begin{tabular}{l|cc|rr}
\toprule
\textbf{Stage} $\rightarrow$ & \multicolumn{2}{c|}{\textbf{Embedding Learning}} & \multicolumn{2}{c}{\textbf{Downstream Task}} \\
\cmidrule(lr){1-1} \cmidrule(lr){2-3} \cmidrule(lr){4-5} 
\textbf{Method} $\downarrow$ & \makebox[1.15 cm][c]{\textbf{NYC}} & \makebox[1.15cm][c]{\textbf{CHI}} & \multicolumn{1}{c}{\textbf{NYC}} & \multicolumn{1}{c}{\textbf{CHI}} \\
\midrule
MVURE      & \textbf{12}   & \textbf{5}    & 0.02 (0.0005) & 0.04 (0.0010) \\
MGFN       & 35   & 42     & \textbf{0.01 (0.0006)} & 0.04 (0.0011) \\
HAFusion   & 26   & 16   & 0.02 (0.0005) & 0.01 (0.0005) \\
ComSRE     & 14   & 11   & 0.02 (0.0006) & \textbf{0.01 (0.0003)} \\
UrbanCLIP  & 506  & 874  & 59 (0.0028)     & 64 (0.0032) \\
FlexiReg   & 67   & 107  & 101 (0.0036)  & 76 (0.0031) \\
HREP       & 16   & 18   & 63 (0.0014)   & 99 (0.0023) \\
ToPT       & 47   & 35   & 149 (0.0005)    & 131 (0.0006) \\
EASTAR     & 105  & 43   & 20 (0.0006)   & 22 (0.0005) \\
\bottomrule
\end{tabular}
} 
\end{table}




\subsection{Ablation Study (Q2)}
We evaluate the contribution of EASTAR’s main design choices through nine ablation variants, organized into four groups. The first group evaluates the channel decomposition design: \textbf{EASTAR-w/o-SB} removes the semantic branch, \textbf{EASTAR-w/o-GB} removes the global branch, and \textbf{EASTAR-w/o-CD} treats event semantics as an additional view without channel decomposition. The second group studies semantic enhancement: \textbf{EASTAR-w/o-SC} removes flow-guided semantic completion, and \textbf{EASTAR-w/o-SGU} removes semantic-guided mobility update. The third group evaluates task adaptation: \textbf{EASTAR-w/o-S2} directly uses Stage-1 representations for downstream prediction, \textbf{EASTAR-w/o-LP} removes LLM-based prior initialization, and \textbf{EASTAR-w/o-AA} replaces adaptive routing with uniform channel fusion. The final variant, \textbf{EASTAR-w/o-CC}, removes the channel constraints used to enforce semantic alignment and channel disentanglement.

\begin{figure}[htbp]
    \centering
    \includegraphics[width=0.95\linewidth]{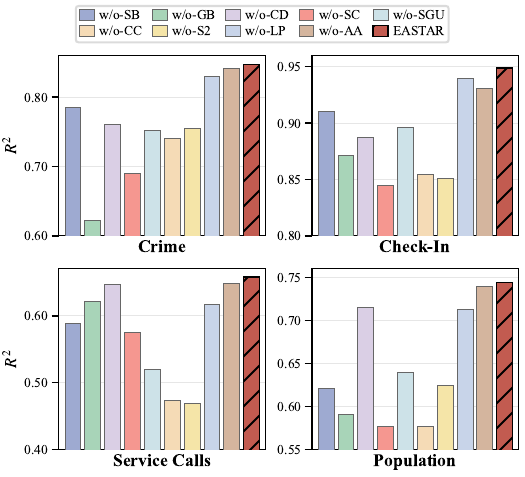}
    \vspace{-3 mm}
    \caption{Ablation study results ($\mathbf{R}^2$ on NYC).}
    \label{fig:ablation}
    \vspace{-3 mm}
\end{figure}

Fig.~\ref{fig:ablation} reports the ablation results on NYC. EASTAR consistently outperforms all variants, showing that each group of designs contributes to the final performance. 
First, representation-structure variants show clear degradation. In particular, EASTAR-w/o-CD remains below the full model, indicating that simply introducing event semantics as an additional view cannot replace event-anchored channel decomposition. EASTAR-w/o-SB and EASTAR-w/o-GB further show that both branches contribute to representation learning.

Second, removing semantic enhancement also hurts performance. EASTAR-w/o-SC indicates that raw event signals are insufficient when used directly, while EASTAR-w/o-SGU shows that injecting completed semantic information back into mobility representations improves the structural branch. Third, task-adaptation variants perform worse than EASTAR, demonstrating that downstream tasks require different channel mixtures. EASTAR-w/o-S2 shows that directly using Stage-1 representations is less effective, the gap of EASTAR-w/o-AA highlights the importance of adaptive routing, and EASTAR-w/o-LP shows that LLM-derived priors provide useful initialization but are not the sole source of task adaptation. Finally, EASTAR-w/o-CC degrades performance, suggesting that semantic alignment and diversity constraints are important for learning coherent and non-redundant channels.

\subsection{Channel Decomposition and Interpretability Analysis (Q3\&Q4)}
In this section, we investigate whether EASTAR's superior performance stems from our architectural design and anchor selection, or merely from accessing new data sources.
\begin{table}[htbp]
\caption{Impact of incorporating Meetup semantics ($\mathbf{R}^2\uparrow$ on NYC)}
\vspace{-2 mm}
\label{tab:model_comparison}
\centering
\footnotesize
\begin{tabular}{l|cccc}
\toprule
\textbf{Variant} & Crime & Check-in & Service Call & Population \\
\midrule
HAFusion        & 0.734  & 0.844  & 0.493  & 0.616 \\
HAFusion$_M$    & 0.744 & 0.857 & 0.505 & 0.620 \\
\midrule
MGFN           & 0.630  & 0.690  & 0.303   & 0.509 \\
MGFN$_M$       & 0.632  & 0.652  & 0.407   & 0.488 \\
\midrule
HREP           & 0.681  & 0.700  & 0.398   & 0.571 \\
HREP$_M$       & 0.660  & 0.667  & 0.368   & 0.536 \\
\midrule
EASTAR-w/o-CD   & 0.761 & 0.887 & 0.648 & 0.715 \\
EASTAR          & \textbf{0.847} & \textbf{0.949} & \textbf{0.658} & \textbf{0.744} \\
\bottomrule
\end{tabular}
\end{table}
\begin{figure*}[htbp]
    \centering
    \includegraphics[width=0.95\linewidth]{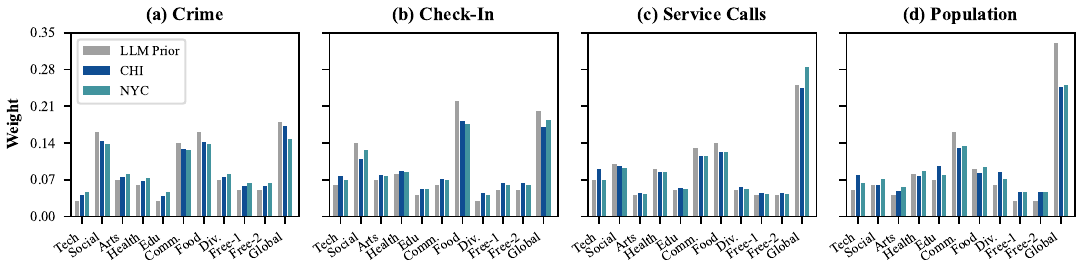}
    \vspace{-5 mm}
    \caption{Comparison between initial LLM-derived priors and the final learned routing weights across downstream tasks.}
    \label{fig:channel_weight}
    \vspace{-3 mm}
\end{figure*}

\textbf{Necessity of explicit channel decomposition.}
To determine whether EASTAR's gains arise from the Meetup event data itself or from our semantic channel decomposition design, we augment representative baselines with Meetup semantic features, denoted by the subscript $_M$ (e.g., HAFusion$_M$). 

As shown in Table~\ref{tab:model_comparison}, the results are mixed: while HAFusion$_M$ achieves slight improvements, MGFN$_M$ and HREP$_M$ suffer noticeable performance drops. This suggests that simply appending semantic features to structural modalities may introduce noise and feature entanglement.

Our ablation variant EASTAR-w/o-CD, which treats semantic profiles as a plain fourth view without channel decomposition, further validates this observation. Although EASTAR-w/o-CD consistently outperforms the augmented baselines, it remains substantially inferior to the full EASTAR model. These results indicate that the performance gains do not come merely from introducing an additional semantic data source. Instead, explicit semantic channel decomposition is essential for transforming heterogeneous mobility interactions into meaningful and task-relevant representations.

\textbf{Impact of semantic anchor sources.}
To evaluate the impact of semantic probe sources on channel decomposition, we replace the event semantic profiles $\mathbf{E}^c$ with two alternatives: (1) \textbf{EASTAR-POI}, where probes are constructed from POI category distributions via a learnable MLP; and (2) \textbf{EASTAR-Random}, where probes are randomly initialized and optimized end-to-end during Stage 1.

As shown in Table~\ref{tab:semantic_probe}, EASTAR consistently outperforms both alternatives across all tasks, confirming the effectiveness of event-based semantic anchors. EASTAR-POI achieves better results than EASTAR-Random, indicating POI distributions still provide useful functional signals. However, a substantial gap remains between EASTAR-POI and EASTAR. We attribute this gap to a semantic granularity mismatch: POI distributions describe what facilities exist in a region, whereas event semantics more directly reflect what activities people engage in and why they travel. Consequently, event-based probes provide better semantic anchors to disentangle heterogeneous mobility flows and learning task-relevant channel representations.

\begin{table}[htbp]
\caption{Impact of semantic probe sources ($\mathbf{R}^2\uparrow$ on NYC)}
\label{tab:semantic_probe}
\vspace{-2 mm}
\centering
\footnotesize
\begin{tabular}{l|cccc}
\toprule
\textbf{Variant} & Crime & Check-in & Service Call & Population \\
\midrule
EASTAR-POI    & 0.778 & 0.874 & 0.508 & 0.584 \\
EASTAR-Random & 0.731 & 0.768 & 0.435 & 0.519  \\
EASTAR        & \textbf{0.847} & \textbf{0.949} & \textbf{0.658} & \textbf{0.744} \\
\bottomrule
\end{tabular}
\end{table}

\textbf{Task-level channel routing analysis.}
To examine whether the learned channels are interpretable and task-adaptive, we compare the learned channel routing weights with the initial LLM priors across all four tasks. As illustrated in Fig.~\ref{fig:channel_weight}, different tasks activate distinct channel combinations. For instance, Check-in prediction prioritizes Food and Social channels, while Crime prediction relies on the Global channel supplemented by Social dynamics. Similar patterns are observed across both cities.

From the weight distributions, we make two observations:

(1) The Global channel maintains a large weight across all tasks, indicating that structural dependencies provide a strong shared predictive backbone, while semantic channels exhibit task-specific variations. In particular, Check-in and Service Call show more pronounced changes in semantic-channel weights, consistent with their stronger dependence on human activity patterns.

(2) The final routing weights deviate from the initial LLM priors, indicating that the prior serves as an initialization rather than determining the final channel selection. Free channels are also up-weighted for several tasks, suggesting that the data-driven router can exploit latent patterns not explicitly covered by the event anchors. This is consistent with the performance degradation of EASTAR-w/o-LP in Fig.~\ref{fig:ablation}.

\begin{figure}[htbp]
    \centering
    \includegraphics[width=1\linewidth]{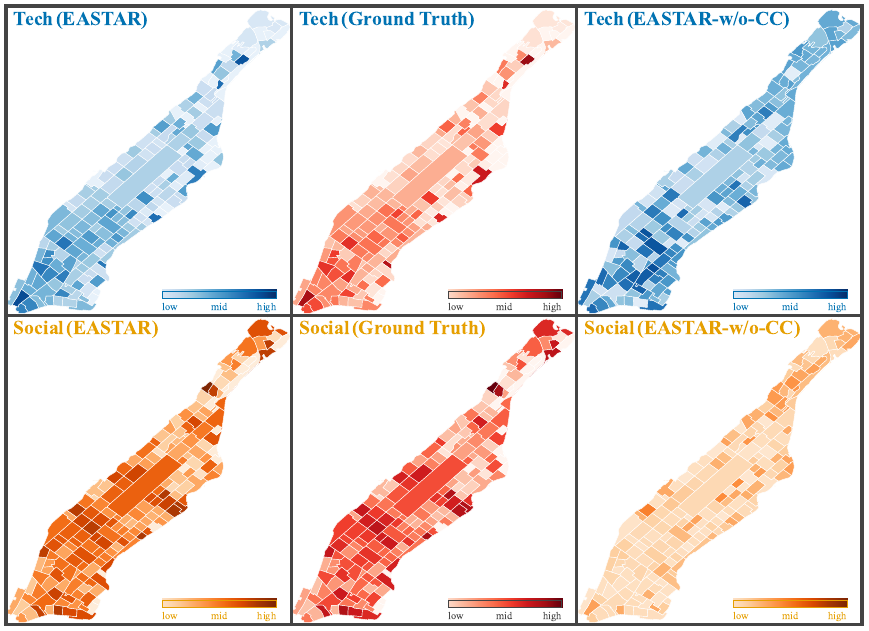}
    \vspace{-7 mm}
    \caption{Region-level semantic consistency on NYC. Activations of the Tech and Social channels from EASTAR and EASTAR-w/o-CC compared with the corresponding ground-truth event distributions.}
    \label{fig:channels}
\end{figure}

\textbf{Region-level spatial verification.}
To examine whether the anchored channels are aligned with observable activity patterns rather than arbitrary latent projections, we compare their activations against the observed event distributions. As shown in Fig.~\ref{fig:channels}, EASTAR’s predicted Tech and Social maps show spatial patterns similar to the observed event distributions. In contrast, the unconstrained variant (EASTAR-w/o-CC) produces diffuse activations that fail to reflect actual event distributions. This provides qualitative evidence for the motivating example in Section I: despite having similar aggregated inflows, EASTAR correctly assigns the Flatiron District a dominant Tech channel and the East Village dominant Food and Social channels, effectively distinguishing regions that are structurally similar but semantically different.

\subsection{Robustness and Sensitivity (Q5)}
We evaluate EASTAR's robustness across different underlying LLMs and its sensitivity to two key hyperparameters.

\textbf{Robustness across different LLMs.}
To verify that our TAPR module does not rely on a specific prior generator, we evaluate EASTAR across four diverse LLMs while keeping the prompt-embedding MLLM fixed: DeepSeek-V4-Flash, Qwen3.5-27B, Gemini-3.5-Flash, and GPT-5.4. As shown in Table~\ref{tab:llm_variants}, EASTAR consistently outperforms the SOTA baselines across all tasks regardless of the chosen LLM. Notably, the model maintains competitive performance even when paired with relatively lightweight open-source models. This suggests that its core effectiveness stems from our architectural design rather than a reliance on massive, proprietary LLMs. While high-capacity models like GPT-5.4 yield the best overall results, the minor performance variance confirms that EASTAR is highly robust and model-agnostic.

\begin{table}[htbp]
\caption{Impact of different LLMs ($\mathbf{R}^2\uparrow$ on NYC)}
\vspace{-2 mm}
\label{tab:llm_variants}
\centering
\footnotesize
\resizebox{\columnwidth}{!}{%
\begin{tabular}{l|cccc}
\toprule
\textbf{Variant} & Crime & Check-in & Service Call & Population \\
\midrule
FlexiReg / ToPT (SOTA)     & 0.751 & 0.899 & 0.497 & 0.701 \\
\midrule
EASTAR-DeepSeek  & 0.832 & 0.941 & 0.606 & 0.705 \\
EASTAR-Qwen      & 0.821 & 0.928 & \textbf{0.661} & 0.710 \\
EASTAR-Gemini    & 0.806 & 0.940 & 0.599 & 0.704 \\
EASTAR-GPT       & \textbf{0.847} & \textbf{0.949} & 0.658 & \textbf{0.744} \\
\bottomrule
\end{tabular}%
}
\end{table}



\begin{figure}[htbp]
    \centering
    \includegraphics[width=1\linewidth]{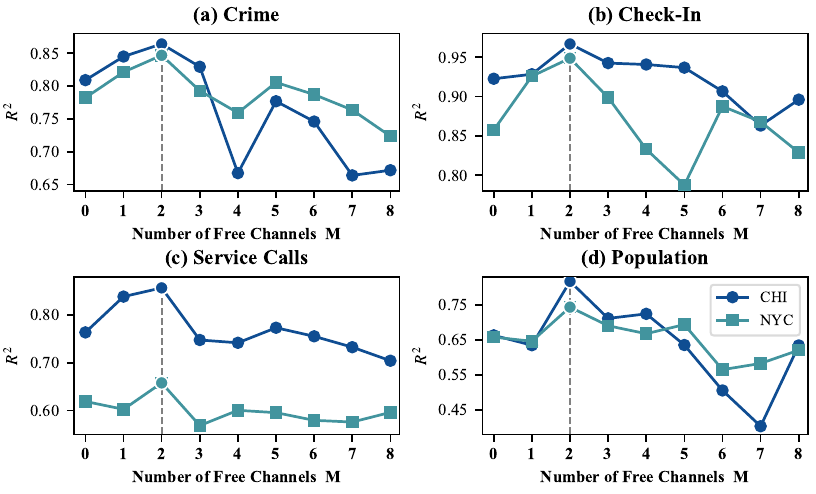}
    \vspace{-6 mm}
    \caption{Impact of the number of free channels $M$.}
    \label{fig:imp of m}
\end{figure}

\textbf{Impact of free channel number $M$.}
We vary the free channel number $M$ from 0 to 8. As shown in Fig.~\ref{fig:imp of m}, performance peaks at $M=2$ across all tasks and then declines. The degradation at $M=0$ confirms that anchored semantic channels alone cannot capture latent mobility patterns absent from event data, validating our free channel design. However, excessive free channels ($M>2$) introduce redundancy and hurt stability, particularly on the smaller CHI dataset. We set $M=2$ for all experiments.

\begin{figure}[htbp]
    \centering
    \includegraphics[width=1\linewidth]{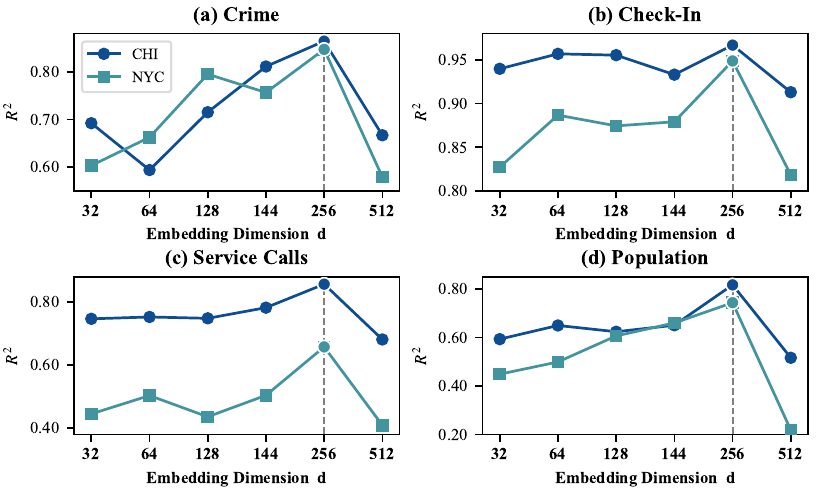}
    \vspace{-6 mm}
    \caption{Impact of embedding dimensionality $d$.}
    \label{fig:imp of d}
\end{figure}

\textbf{Impact of embedding dimensionality $d$.}
We explore the embedding dimensionality $d \in \{32, 64, 128, 144, 256, 512\}$. As shown in Fig.~\ref{fig:imp of d}, accuracy improves steadily up to $d=256$, as higher dimensionality provides sufficient capacity for each channel in the $(K{+}M{+}1)$-channel representation. Performance drops at $d=512$, likely due to overfitting on the limited number of regions. Unlike $M$, which controls the number of free channels, $d$ determines the representation capacity of each channel. We set $d=256$ for all experiments.

\section{Conclusion}
We proposed EASTAR, an event-anchored and task-adaptive framework for urban region representation learning. EASTAR uses offline activity events as sparse semantic anchors to factorize aggregated mobility representations into purpose-related semantic channels, while preserving global spatial structure and latent mobility patterns through global and free channels. It further introduces a task-adaptive prompt router that initializes task-channel relevance from LLM-derived priors and refines the routing weights through downstream supervision. Experiments on two real-world urban datasets across four prediction tasks show that EASTAR consistently outperforms representative baselines, achieving up to 39.15\% reduction in MAE. 
Ablation and interpretability analyses confirm that explicit within-flow channel factorization outperforms naive semantic data augmentation, and that the learned routing patterns are consistent across cities.
Limitations include evaluation on only two cities and reliance on event data as sparse semantic anchors rather than ground-truth trip-purpose labels; future work will study broader urban contexts and richer activity sources.

\section*{Acknowledgment}

This research was supported by the National Natural Science Foundation of China (No.~62306184); the Natural Science Foundation for Top Talents of SZTU (Nos.~GDRC202518 and GDRC202320); the Shenzhen Science and Technology Program (Nos.~RCBS20231211090548077 and JCYJ20240813113218025); the Guangdong Basic and Applied Basic Research Foundation (No.~2026A1515010133); and the Project for Improving Scientific Research Capabilities of Key Construction Disciplines in Guangdong Province (No.~2025ZDJS039).

\section*{AI-Generated Content Acknowledgement}

OpenAI's ChatGPT was used solely to refine language and presentation throughout the manuscript. The authors verified all content and take full responsibility for it.

\bibliographystyle{IEEEtran}
\bibliography{IEEEabrv, myrefs}

\end{document}